\pdfoutput=1

\documentclass[11pt]{article}

\usepackage[]{acl}

\usepackage{times}
\usepackage{latexsym}

\usepackage[T1]{fontenc}

\usepackage[utf8]{inputenc}

\usepackage{microtype}
\usepackage{graphicx}
\usepackage{subfigure}
\usepackage{svg}
\usepackage{CJKutf8}
\usepackage{setspace}
\usepackage{multirow}
\usepackage{subfloat}
\usepackage{graphicx}
\usepackage{url}
\usepackage{tikz}
\usepackage{pgfplots}
\usepackage{color}
\usepackage{mathtools}
\usepackage{makecell}
\usepackage{colortbl,booktabs}
\usepackage{paralist}
\usepackage{amsmath,amssymb,bm}
\usepackage{threeparttable}

\def\So{$S_{open-end}$}
\def\Sno{$S_{non-open-end}$}
\def\Sc{$S_{conversation}$}
\def\Sd{$S_{data-to-text}$}
\def\Sq{$S_{question}$}
\def\Sg{$S_{general}$}

\title{SkillNet-NLG: General-Purpose Natural Language Generation with a Sparsely Activated Approach}

\author{Junwei Liao, Duyu Tang\thanks{ \ \ Contact: Duyu Tang ({duyutang@tencent.com}).}\ , \ Fan Zhang \and Shuming Shi \\ \\  Tencent AI Lab}
\begin{document}
\maketitle
\begin{abstract}
We present SkillNet-NLG, a sparsely activated approach that handles many natural language generation tasks with one model.
Different from traditional dense models that always activate all the parameters, SkillNet-NLG selectively activates relevant parts of the parameters to accomplish a task, where the relevance is controlled by a set of predefined skills.
The strength of such model design is that it provides an opportunity to precisely adapt relevant skills to learn new tasks effectively.
We evaluate on Chinese natural language generation tasks.
Results show that, with only one model file, SkillNet-NLG outperforms previous best performance methods on four of five tasks. 
SkillNet-NLG performs better than two multi-task learning baselines (a dense model and a Mixture-of-Expert model) and achieves comparable performance to task-specific models.
Lastly, SkillNet-NLG surpasses baseline systems when being adapted to new tasks.
\end{abstract}

 \vspace{0.001cm}
\section{Introduction}
The flexibility of Transformer \cite{vaswani2017attention} facilitates the development of multitask models that use one model to handle multiple tasks~\cite{liu2019multi,raffel2019exploring,lewis-etal-2020-bart}.
These models are typically ``dense'' \textemdash \  all the model parameters are activated for all the tasks. 
However, it is unclear what skills are learned in which parts of the parameters.
Even though tackling different tasks requires different skills  \cite{dean-pathways-2021,tang2022skillnet}, dense models do not allow us to carry out subtle operations to choose different skills for different tasks.
Moreover, when adapting a well-trained dense model to learn new tasks, all the encoded ``vague'' skills are transferred blindly, regardless of their relevance to the tasks.

\begin{figure}[!t]
	\centering
	\includegraphics[width=1\linewidth]{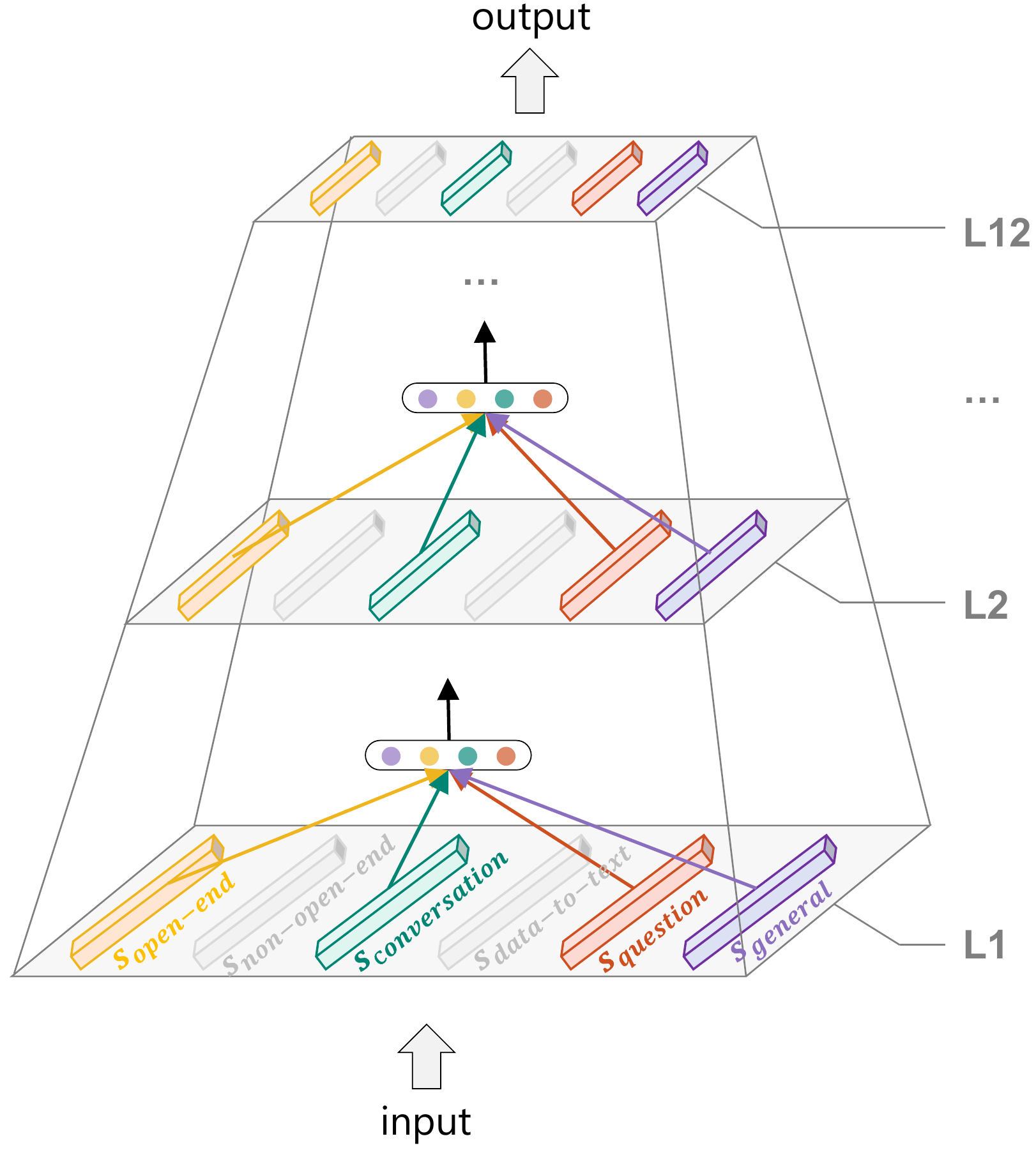}
	\caption{An illustration of our sparsely activated model SkillNet-NLG for dialogue generation. Each pillar represents a skill module and pillars filled in color (e.g., yellow, green, red and purple) are activated.}
	\label{fig:model-intro}
\end{figure}

\begin{table}[htbp]
  \centering
  \resizebox{\linewidth}{!}{
    \begin{tabular}{ll}
    \toprule
    \textbf{Skill} & \textbf{Definition} \\
    \midrule
    \So    & open-ended text generation \\
    \Sno   & non-open-ended text generation \\
    \Sc    & understand the conversational contexts \\
    \Sd    & generate text from structured data \\
    \Sq    & understand natural language questions \\
    \Sg    & generic skill \\
    \bottomrule
    \end{tabular}   }
  \caption{Skills and definitions of SkillNet-NLG.}
  \label{tab:skills_definition}
\end{table}

In this work, we propose a general-purpose natural language generation model called SkillNet-NLG.
The basic idea is that the model includes multiple skill modules, each of which stands for a particular skill defined in Table~\ref{tab:skills_definition}.
Instead of activating all the parameters in traditional dense models, we only activate relevant skills for a downstream task.
As depicted in Figure~\ref{fig:model-intro}, for the task of dialogue generation, SkillNet-NLG requires the ability to generate open-ended language (\So), understand the conversational context (\Sc) and understand natural language questions (\Sq).
Therefore, the skill modules related to \So, \Sc, \Sq~and \Sg\footnote{We define a  general skill \Sg, which works as a default skill and is always activated.} are activated.
The remaining modules (\Sno~and \Sd) are not activated.
We develop SkillNet-NLG based on Transformer~\cite{vaswani2017attention} with an encoder-decoder structure.
We modify every other layer in both Transformer encoder and decoder through replacing one FFN layer with multiple FFN layers, each of which corresponds to a skill.

We conduct extensive experiments on Chinese natural language generation tasks\footnote{Our approach is language agnostic. We leave the extension of SkillNet-NLG to more languages in the future.}.
We consider five tasks (text summarization, advertisement generation, question answering, dialogue generation and grammatical error correction) for multi-task training.
With one model, SkillNet-NLG outperforms previous best performance methods on four of five tasks and performs better than both dense and MoE multitask baselines. 
We further adapt well-trained models to three new tasks (topic-to-essay generation, paraphrase generation and story generation), and find that SkillNet-NLG performs better than all the aforementioned baseline systems.

\section{Methods}
In this section, we first briefly introduce the Transformer (\S\ref{ssec:transformer}), which we use as the backbone of SkillNet-NLG. Then, we describe the proposed model architecture (\S\ref{ssec:skillnet-nlg}). Finally, we present how to do multi-task training (\S\ref{ssec:model_training}) with SkillNet-NLG.

\subsection{Transformer}
\label{ssec:transformer}

Transformer~\cite{vaswani2017attention} model has an encoder-decoder structure with multiple layers.
In the encoder, each layer includes a multi-head self-attention network (Attention) and a feed forward network (FFN) layer.
Specifically, given the layer input $\boldsymbol{h}_{in}$, the layer output is computed as follows,
\begin{equation}
\boldsymbol{h}_{out}=\mathrm{FNN}(\mathrm{Attention}(\boldsymbol{h}_{in})).
\end{equation}

Each layer of the decoder is similar with the encoder except that it inserts an additional Attention layer, which performs multi-head attention over the output of the encoder stack.
Since Transformer is a commonly used model in natural language processing, we exclude a detailed description and refer readers to the original paper.

\subsection{SkillNet-NLG}
\label{ssec:skillnet-nlg}
We develop SkillNet-NLG using Transformer~\cite{vaswani2017attention} as the backbone.
As shown in Figure~\ref{fig:model-intro}, the high level idea is that SkillNet-NLG has multiple skill modules and only activates relevant skills when it is adopted to a downstream task.
Specifically, we modify a Transformer layer  (for both encoder and decoder) through replacing one FFN layer with multiple FFN layers, each of which corresponds to a skill.  
When the model handles a task, only the FFN layers corresponding to relevant skills are activated.
For example, for the task of dialogue generation, we only activate \So, \Sc, \Sq~and \Sg.
The remaining modules (\Sno~and \Sd) are not activated.
For a particular FFN layer $\mathrm{FFN}_{k}$, it works same with the original FFN layer and produces 
skill-specific representations as follows,
\begin{equation}
\boldsymbol{h}_{k}=\mathrm{FFN}_{k}(\mathrm{Attention}(\boldsymbol{h}_{in})).
\end{equation}
Since the size of the set of activated modules is variable, we compute the output representations using the average pooling as follows,
\begin{equation}
\boldsymbol{h}_{out}=\frac{1}{|S|}\sum_{k=1}^{|S|} \boldsymbol{h}_{k},
\end{equation}
where $S$ is the set of activated skills.
For the the task of dialogue generation, as shown in Figure~\ref{fig:model-intro}, $S = \{\text{\So, \Sc, \Sq, \Sg}\}$.
The remaining operations in SkillNet-NLG are same as the original Transformer.
Following \citet{lepikhin2020gshard}, we only make the above changes in every other Transformer layer to avoid adding too many parameters.

\begin{table*}[!t]
\centering
\resizebox{\textwidth}{!}{
\begin{tabular}{lcccccc}
\toprule
\multicolumn{1}{c}{\multirow{2}[4]{*}{\textbf{Task}}} & \multicolumn{6}{c}{\textbf{Skills}} \\
\cmidrule{2-7}      & \multicolumn{1}{l}{\So} & \multicolumn{1}{l}{\Sno} & \multicolumn{1}{l}{\Sc} & \multicolumn{1}{l}{\Sd} & \multicolumn{1}{l}{\Sq} & \multicolumn{1}{l}{\Sg} \\
\midrule
\multicolumn{7}{l}{\textit{Tasks for training the multi-task models }} \\
\midrule
Text Summarization &       & \checkmark &       &       &       & \checkmark \\
Advertisement Generation & \checkmark &       &       & \checkmark &       & \checkmark \\
Question Answering & \checkmark &       &       &       & \checkmark & \checkmark \\
Dialogue Generation & \checkmark &       & \checkmark &       & \checkmark & \checkmark \\
Grammatical Error Correction &       & \checkmark &       &       &       & \checkmark \\
\midrule
\multicolumn{7}{l}{\textit{New tasks for fine-tuning well-trained multi-task models}} \\
\midrule
Topic-to-Essay Generation & \checkmark &       &       & \checkmark &       & \checkmark \\
Paraphrase Generation &       & \checkmark &       &       &       & \checkmark \\
Story Generation & \checkmark &       &       &       &       & \checkmark \\
\bottomrule
\end{tabular}%
}
\caption{
Relations between tasks and skills. Relevant skills for each task are marked with ticks.
}
\label{tab:task_skills}
\end{table*}

\subsection{Model Training}
\label{ssec:model_training}
 
The model is trained on the mixing of training samples from all tasks. 
In each iteration, a mini-batch is selected from one task.
A task-specific prefix is appended to the input. The model computes the cross-entropy loss between the generated text and the reference text to update the model parameters.
Since the training data of different tasks are unbalanced, we follow \citet{tang2022skillnet} and 
adopt a temperature-scaled mixing strategy for data sampling.
Specifically, we sample mini-batches from $N$ tasks according to probability $\{p_1, \ldots, p_N\}$:
\begin{equation}
p_i = \frac{D_i^{\frac{1}{T}}}{\sum_{j=1}^N D_j^{\frac{1}{T}}} \ \ \text{with} \ \ D_i = \min(n_i, K),
\end{equation}
where $n_i$ is the number of training examples for the $i$-th task.
$K$ is a hyper parameter.
$T$ is the sampling temperature. The distribution is equivalent to original data distribution for $T = 1$ and is close to the uniform distribution for larger value (e.g., $T = 1024$).
We analyze the influence of $T$ in \S \ref{sec:analysis-T}.

\begin{table*}[!t]
  \centering
    \begin{tabular}{l|ccccc|c}
    \toprule
          & \textbf{LCSTS} & \textbf{AdGen} & \textbf{MATINF-QA} & \textbf{KdConv} & \textbf{NLPCC} & \textbf{Avg} \\
    \midrule
    Previous best system  & 41.87{$^\dagger$}  & 10.63{$^\dagger$}  & 20.51{$^\dagger$}  & 18.50{$^\ddagger$}  & \textbf{36.97}{$^*$} & 25.70  \\
    \midrule
    Task-specific fine-tuning & 42.05  & 10.38  & \textbf{21.06} & 21.11  & 36.42  & 26.20  \\
    Joint fine-tuning (Dense) & 41.77  & 10.25  & 20.32  & \textbf{21.16} & 36.19  & 25.94  \\
    Joint fine-tuning (MoE) & 41.80  & 10.25  & 20.56  & 20.71  & 35.96  & 25.86  \\
    \midrule
    SkillNet-NLG & \textbf{42.40} & \textbf{10.80} & 20.73  & 20.76  & 36.68  & \textbf{26.27} \\
    \bottomrule
    \end{tabular}%
  \caption{Test results on the five task datasets during multi-task training. \textbf{Avg} is the average score of all tasks. $^\dagger$~indicates the score from CPT-Large \cite{shao2021cpt}. $^\ddagger$ indicates the score from mBART-Large \cite{liu2020multilingual}. $^*$ indicates the score from Mask GEC \cite{zhao2020maskgec}.
  }
  \label{tab:results}%
\end{table*}%

\section{Experiments}
In this section, we describe experiment settings and report results.

\subsection{Experimental Setup}
\label{sec:experiment-setup}

We consider five tasks for multi-task training. 
We compare with the following baselines.

$\bullet$ \textbf{Task-specific fine-tuning}: We fine-tune all the parameters of our BART model\footnote{We pre-train a strong Chinese BART model on a collection of 800G of web news data.} for each task individually. As a result, we get a total of five task-specific models for five tasks.

$\bullet$ \textbf{Joint fine-tuning (Dense)}: We fine-tune the BART model jointly on five tasks.

$\bullet$ \textbf{Joint fine-tuning (MoE)}:
We train a Mixture-of-Experts (MoE) baseline \cite{lepikhin2020gshard} with the same amount of six experts. 
For each token, we use a gating function to selectively activate the top-2 experts.
The parameters of the model are initialized with our BART model and learned jointly on five tasks.

Table~\ref{tab:task_skills} presents these tasks and the activated skills for each task.
Following existing works, we report ROUGE-L for LCSTS and MATINF-QA datasets, BLEU-4 for AdGen and KdConv datasets, $F_{0.5}$ for NLPCC dataset, respectively. 
We average these scores as a reference to the overall performance. 
Dataset statistics and training details are presented in Appendix~\ref{sec:datasets} and~\ref{sec:train_para}, respectively.

\subsection{Overall Results}
\label{sec:experiment-results}

Table~\ref{tab:results} shows the results of the baselines as well as SkillNet-NLG on five tasks.
Overall, SkillNet-NLG performs better than task-specific fine-tuning and two multi-task learning baselines (i.e., Joint fine-tuning (Dense) and Joint fine-tuning (MoE)) in terms of the average score.
With only one model, SkillNet-NLG outperforms previous best methods on four of five tasks, demonstrating the effectiveness of the sparsely activated approach.

\subsection{Adaptation to New Tasks}

In this section, we adapt models that are well-trained on five tasks to new tasks separately.

\begin{table}[!h]
  \centering
  \resizebox{\linewidth}{!}{  
    \begin{tabular}{lccc}
    \toprule
          & \textbf{ZhiHu} & \textbf{PKUPB} & \textbf{OutGen} \\
    \midrule
    Previous best system  & \textbf{11.02}$^\dagger$ & --    & 24.77$^\ddagger$  \\
    \midrule
    Task-specific fine-tuning & 10.56  & 31.88  & 25.23  \\
    Joint fine-tuning (Dense) & 10.53  & 31.93  & 24.47  \\
    Joint fine-tuning (MoE) & 10.83  & 31.51  & 24.23  \\
    \midrule
    SkillNet-NLG & 10.98  & \textbf{32.02 } & \textbf{25.99 } \\
    \bottomrule
    \end{tabular}%
  }
  \caption{Test results on three new task datasets. Results with $^\dagger$ are from SCTKG(Gold-Senti) \cite{qiao2020sentiment}. Results with $^\ddagger$ are from LongLM$_{large}$ \cite{guan2021lot}.}
  \label{tab:new_task_results}%
\end{table}%

Table~\ref{tab:new_task_results} shows the results of different models on three new tasks. Following existing studies, we report BLEU-2 for ZhiHu and OutGen datasets and report BLEU-4 for the PKUPB dataset.
We can see that SkillNet-NLG outperforms task-specific fine-tuning and two multi-task baselines.
SkillNet-NLG achieves comparable performance with~\citet{qiao2020sentiment} on ZhiHu, which uses external knowledge base.
SkillNet-NLG achieves a 1.22 improvement compared to the LongLM$_{large}$, which has larger number (i.e.,  one billion) of parameters and is pre-trained on a large-scale in-domain data.

\subsection{Influence of Data Sampling Strategies}\label{sec:analysis-T}
\begin{figure}[!h]
	\centering
	\includegraphics[width=1\linewidth]{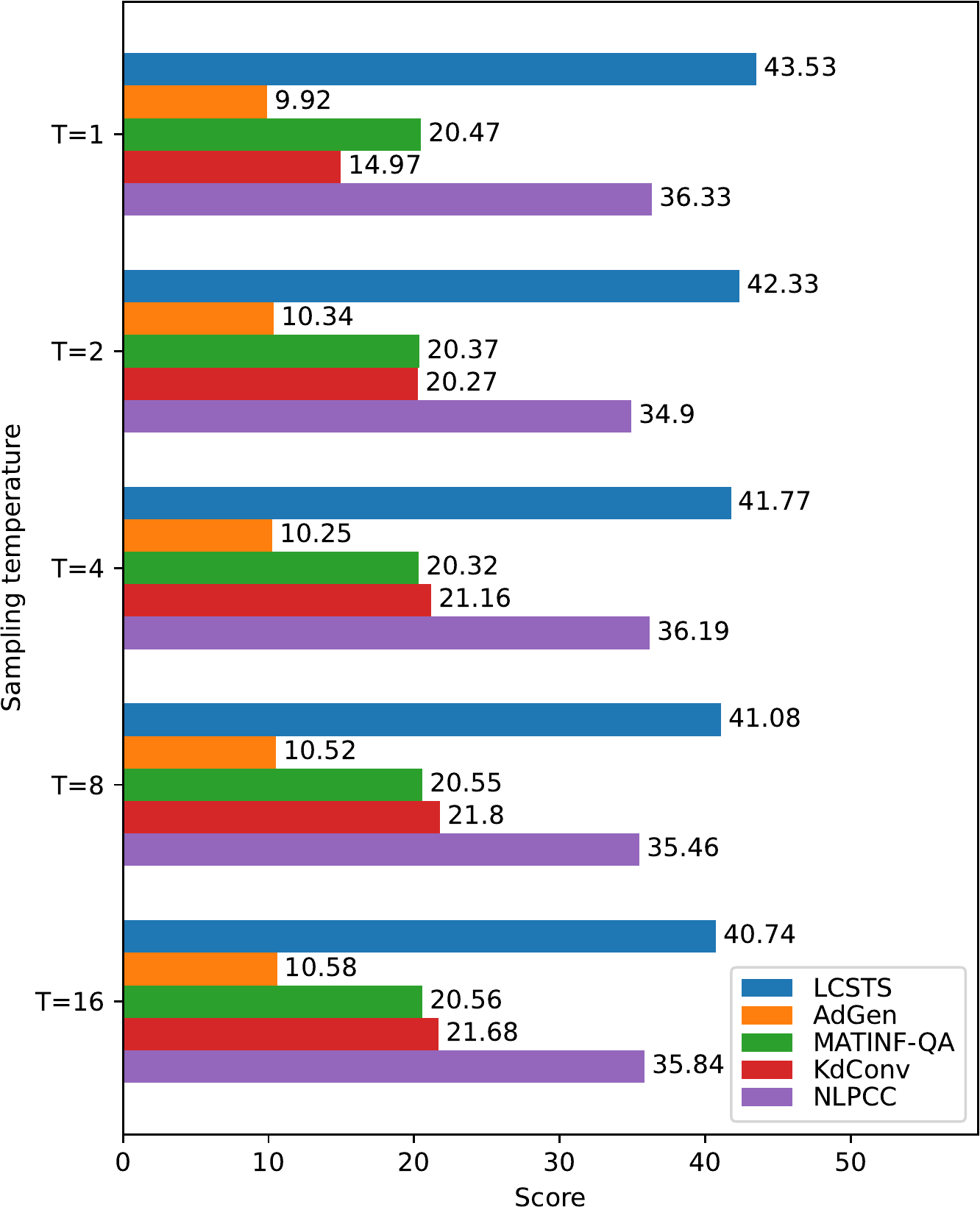}
	\caption{Results on the development sets of five tasks with different data sampling strategies.}
	\label{fig:sampling_temperature}
\end{figure}

As described in Section~\ref{ssec:model_training}, we sample training examples from each task by changing the sampling temperature $T$. Figure~\ref{fig:sampling_temperature} shows the scores with different values of $T$ on the development sets of the five tasks. 
When $T=1$, the training examples are sampled in proportion to the size of each task's training data. Since these data sets are very unbalanced (as given in Table \ref{tab:task_dataset}), the high-resource task of LCSTS gets the highest score while the low-resource task of KdConv gets the lowest score.
As $T$ increases, the data imbalance between high-resource and low-resource tasks gradually decreases.
When $T=4$, the model reaches a balance between two extremes and achieves the best average score on the development sets. Therefore, we adopt $T=4$ throughout all experiments.

\section{Conclusion}
In this work, we present a general-purpose model called SkillNet-NLG.
It deals with multiple natural language generation tasks with one model.
The key feature of our approach is that it is sparsely activated guided by a set of predefined skills. Only the parameters of relevant skills are activated. The advantage of such model design is that it enables us to only transfer relevant skills to learn new tasks.
Experimental results on Chinese NLG tasks verify the effectiveness of our approach.
In the future, we plan to adapt the model to more languages and even more modalities.

\bibliography{anthology,custom,acl}
\bibliographystyle{acl_natbib}

\appendix

\section{Computational Cost}
\label{sec:compute_cost}

Table~\ref{tab:para_num} shows the number of parameters of models.
The number of activated parameter of SkillNet-NLG depends on the number of activated skill modules when performing a specific task (Table~\ref{tab:para_num}).

\begin{table*}[!t]
  \centering
    \begin{tabular}{cccccc}
    \toprule
    \textbf{Task} & \textbf{Dataset} & \textbf{\#Train} & \textbf{\#Dev} & \textbf{\#Test} & \textbf{Metrics} \\
    \midrule
    \multicolumn{6}{l}{\textit{Tasks for training the multi-task models }} \\
    \midrule
    Text Summarization & LCSTS & 2160k & 30k   & 725   & ROUGE-L \\
    Advertisement Generation & AdGen & 114k  & 1k    & 3k    & BLEU-4 \\
    Question Answering & MATINF-QA & 740k  & 100k  & 210k  & ROUGE-L \\
    Dialogue Generation & KdConv & 63k   & 9k    & 9k    & BLEU-4 \\
    Grammatical Error Correction & NLPCC & 1200k & 5k    & 2k    & $F_{0.5}$ \\
    \midrule
    \multicolumn{6}{l}{\textit{New tasks for fine-tuning well-trained multi-task models}} \\
    \midrule
    Topic-to-Essay Generation & ZhiHu & 27k   & 300   & 2.3k  & BLEU-2 \\
    Paraphrase Generation & PKUPB & 490k  & 10k   & 10k   & BLEU-4 \\
    Story Generation & OutGen & 1456  & 242   & 729   & BLEU-2 \\
    \bottomrule
    \end{tabular}%
  \caption{Statistic of datasets.
  }
  \label{tab:task_dataset}
\end{table*}

\begin{table*}[htbp]
  \centering
\resizebox{\linewidth}{!}{
\begin{tabular}{lclcc}
\toprule
      & \textbf{\#Total Params} & \textbf{Task} & \textbf{\#Skill} & \textbf{\#Params Activated} \\
\midrule
\multicolumn{1}{l}{Task-specific fine-tuning} & 376.46M & ---   & ---   & 376.46M \\
\multicolumn{1}{l}{Joint fine-tuning (Dense)} & 376.46M & ---   & ---   & 376.46M \\
\multicolumn{1}{l}{Joint fine-tuning (MoE)} & 880.28M & ---   & ---   & 477.28M \\
\midrule
\multicolumn{1}{l}{\multirow{8}[2]{*}{SkillNet-NLG}} & \multirow{8}[2]{*}{880.20M} & Text Summarization & 2     & 477.20M \\
      &       & Advertisement Generation & 3     & 577.95M \\
      &       & Question Answering & 3     & 577.95M \\
      &       & Dialogue Generation & 4     & 678.70M \\
      &       & Grammatical Error Correction & 2     & 477.20M \\
      &       & Topic-to-Essay Generation & 3     & 577.95M \\
      &       & Paraphrase Generation & 2     & 477.20M \\
      &       & Story Generation & 2     & 477.20M \\
\bottomrule
\end{tabular}%
}
  \caption{The number of parameters of models.}    
  \label{tab:para_num}%
\end{table*}%

\section{Datasets}
\label{sec:datasets}

Table~\ref{tab:task_dataset} shows statistic of all Chinese datasets used in experiments.
We first use five task datasets to train multi-task models and evaluate the performance. 
Then we use another three task datasets to fine-tune the models respectively.

\textbf{Text summarization} is designed to facilitate a quick grasp of the essence of an input document by producing a condensed summary of its content. \textbf{LCSTS} is a large scale Chinese short text summarization dataset \cite{hu-etal-2015-lcsts} collected from Sina Weibo. We use the same data division and evaluation metric as~\citet{shao2021cpt}.

\textbf{Advertisement Generation} aims to generate a long advertisement given a set of attribute value pairs of a commodity. \textbf{AdGen} consists of 119K pairs of clothing specification tables and their advertising texts from a Chinese e-commerce platform. Following \citet{shao2021cpt}, We use the same data pre-processing and format the input data as a list of attribute value pairs.

\textbf{Question answering} is to produce an answer in natural language given a question. \textbf{MATINF-QA} is a large-scale Chinese Open Domain QA dataset collected by~\citet{xu-etal-2020-matinf}, which contains 1.07 million question-answer pairs from the health domain. maternity and baby caring

\textbf{Dialogue generation} is to generate a response based on historical utterances in a dialogue. \textbf{KdConv} is a multi-domain knowledge-driven conversation dataset containing 4.5K conversations from three domains \cite{zhou-etal-2020-kdconv}. We follow \citet{sun2021ernie} for data splitting and pre-processing, and exclude knowledge triplets from the input.

\textbf{Grammatical Error Correction (GEC)} is the task of correcting different kinds of errors in text such as spelling, punctuation, grammatical, and word choice errors. \textbf{NLPCC} provided by NLPCC 2018 Shared Task\footnote{http://tcci.ccf.org.cn/conference/2018/taskdata.php} \cite{zhao2018overview} contains large-scale Chinese texts written by non-native speakers in which grammatical errors have been annotated and corrected by native speakers. We use the official MaxMatch (${M}^2$) scorer to evaluate models\footnote{http://www.comp.nus.edu.sg/nlp/software.html}.

\textbf{Topic-to-essay} takes a set of topic words as input and outputs an essay (a paragraph) under the theme of the topics. 
\textbf{ZhiHu} is a topic-to-essay dataset \cite{feng2018topic} crawled from ZhiHu, a Chinese question-and-answer website. It consists of 100 high frequent topic words and Chinese essays whose length is between 50 and 100. We use the same data split and evaluation metric as \citet{yang2019enhancing}\footnote{The dataset can be download by https://pan.baidu.com/s/17pcfWUuQTbcbniT0tBdwFQ}.

\textbf{Paraphrase generation} is the task of generating an output sentence that preserves the meaning of the input sentence but contains variations in word choice and grammar.
\textbf{PKU Paraphrase Bank (PKUPB)} is a large-scale sentence-level paraphrase corpus for Chinese that contains 509,832 sentence pairs\footnote{https://github.com/pkucoli/PKU-Paraphrase-Bank}. We randomly sample 10,000 pairs as the validation and test set respectively and use the remaining part as training set.

\textbf{Story generation} aims to generating a reasonable story from a leading context. The story must remain thematically consistent across the complete document as well as keeping creativity. 
\textbf{OutGen} is an outline-conditioned story generation dataset introduced by \citet{guan2021lot}, which requires generating a coherent long-form story conditioned on an outline of characters and events. The outline is a set of out-of-order phrases. We use the same data split and evaluation metrics provided by \citet{guan2021lot}\footnote{The data and evaluation script are available at https://github.com/thu-coai/LOT-LongLM.}.

\section{Model Training}
\label{sec:train_para}

\subsection{Multitask Training}

We build our SkillNet model using the implementation of BART-large by HuggingFace’s Transformers\footnote{https://github.com/huggingface/transformers} \cite{wolf2020huggingfaces}, which has 12 encoder layers, 12 decoder layers, 1024 hidden state dimensions and 4096 FFN dimensions. All the skill modules are initialized with FFN layers from our pre-trained Chinese BART. 
We conduct multi-task training for 100k steps with maximum source length of 512, maximum target length of 200 and batch size of 512. We use Adam \cite{kingma2014adam} as the optimizer with $\beta_1 = 0.9, \beta_2 = 0.999, \epsilon = 1e^{-8}$. The learning rate is warmed up over the first 10k steps to a peak value of $3e^{-5}$, and then linearly decayed. We show the learning curve of each task in Appendix~\ref{sec:learning_curves}. 
We set the size limit $K=2^{21}$ and the sampling temperature $T=4$ after searching in $\{1,2,4,8,16,1024\}$.
In inference stage, we use the beam search decoding and set the beam size to 4 for all tasks.

\subsection{New Tasks Training}

Table~\ref{tab:new_task_para} shows the specific hyper-parameters used to train three new tasks. Other training parameters are the same as for multitask training.

\begin{table}[htbp]
  \centering
  \resizebox{\linewidth}{!}{
    \begin{tabular}{lccc}
    \toprule
          & \textbf{ZhiHu} & \textbf{PKUPB} & \textbf{OutGen} \\
    \midrule
    Epochs & 16    & 6     & 16 \\
    Batch size & 128   & 64    & 64 \\
    Learning rate & 3e-5  & 3e-5  & 5e-5 \\
    Max source length & 30    & 140   & 100 \\
    Max target length & 170   & 140   & 310 \\
    Metric for best model & BLEU-2 & BLEU-4 & BLEU-2 \\
    \bottomrule
    \end{tabular}%
  }
  \caption{Training parameters for fine-tuning well-trained SkillNet-NLG on new tasks.}
  \label{tab:new_task_para}%
\end{table}%

\section{Learning Curves}
\label{sec:learning_curves}

\begin{figure*}[!t]
	\centering
	\subfigure[Training loss on all task datasets.]{
	\includegraphics[width=0.32\linewidth]{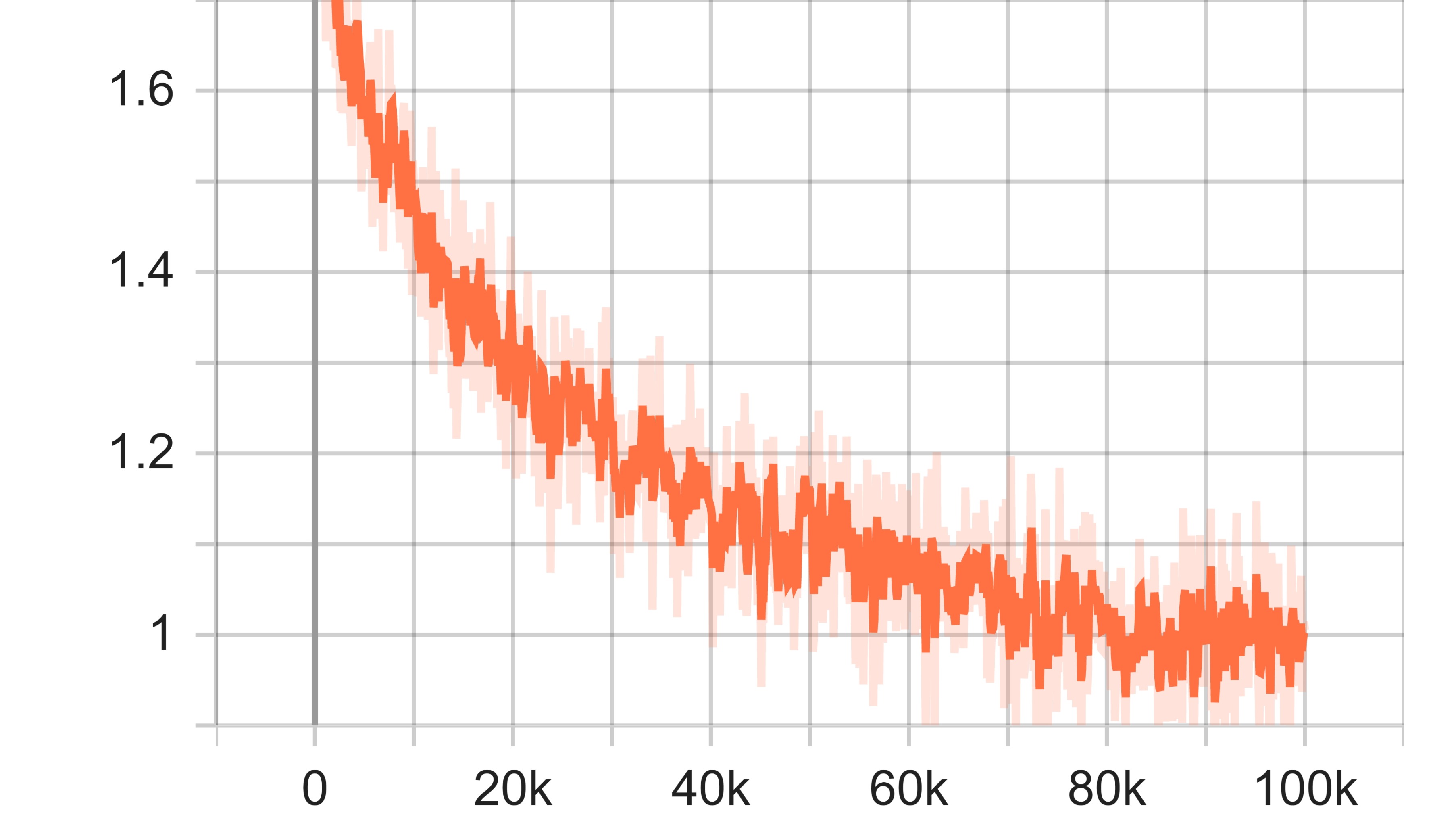}}
	\subfigure[Training loss on LCSTS.]{
	\includegraphics[width=0.32\linewidth]{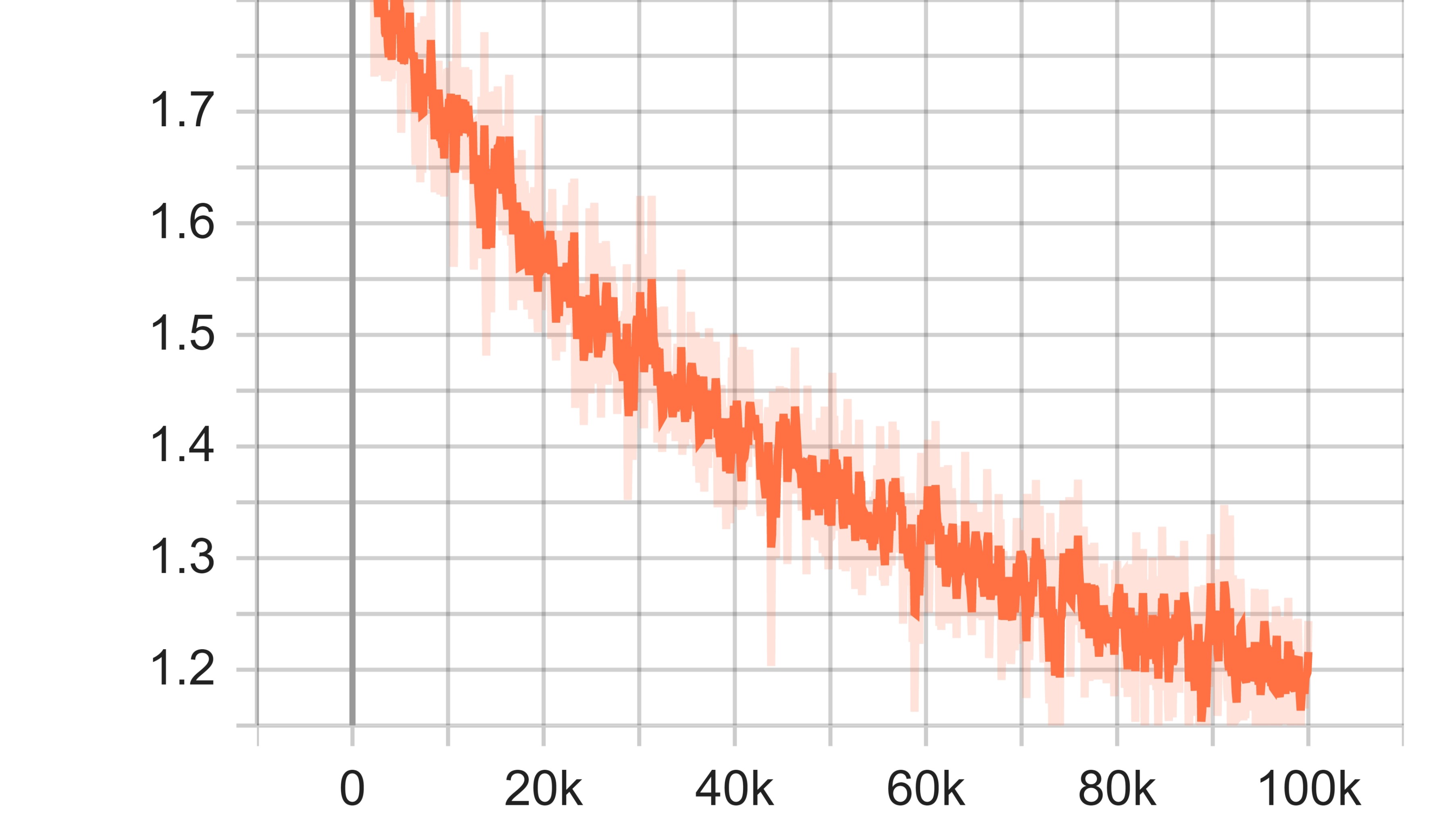}}
	\subfigure[Training loss on AdGen.]{
	\includegraphics[width=0.32\linewidth]{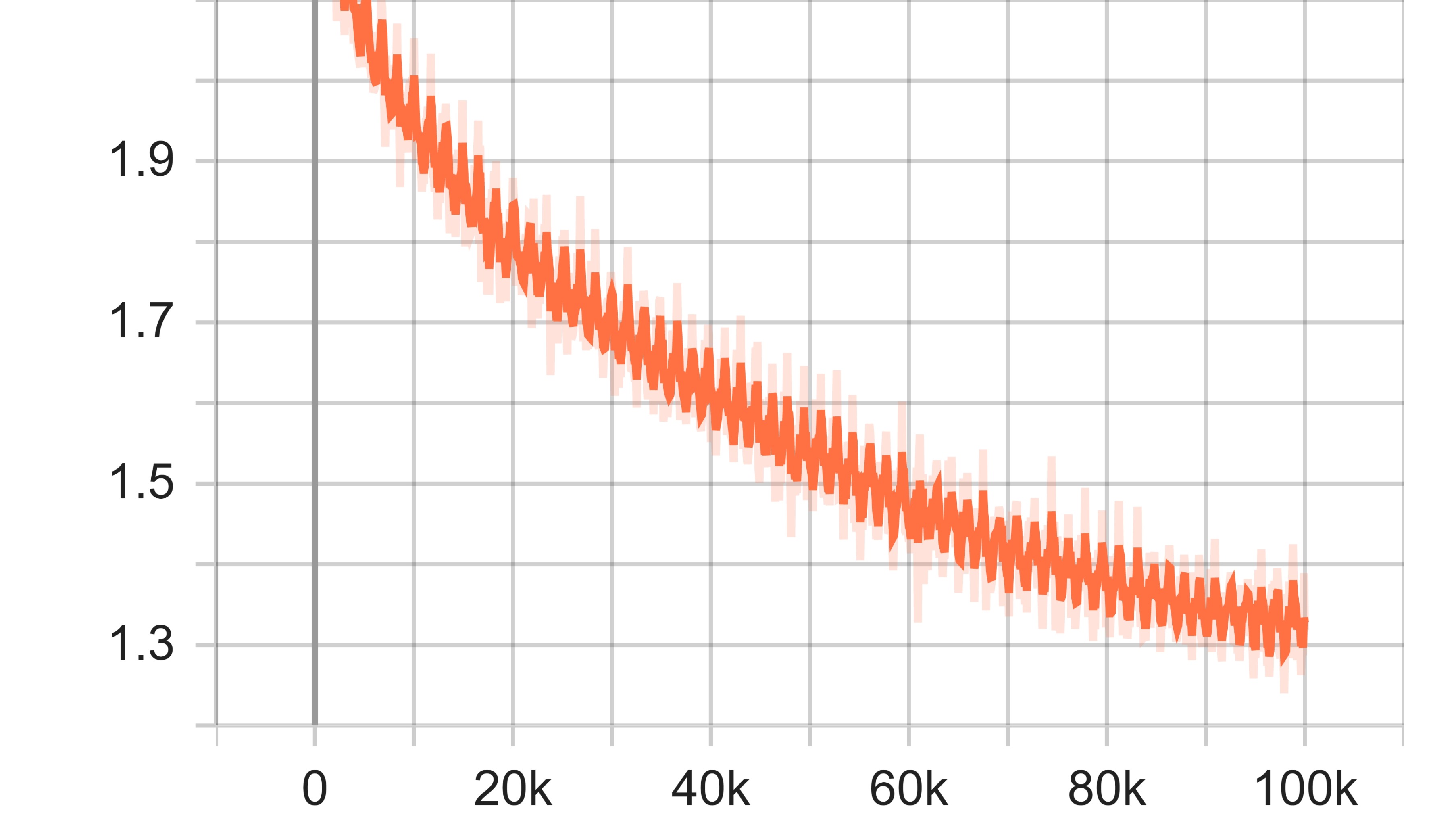}}
	
	\subfigure[Training loss on MATINF-QA.]{
	\includegraphics[width=0.32\linewidth]{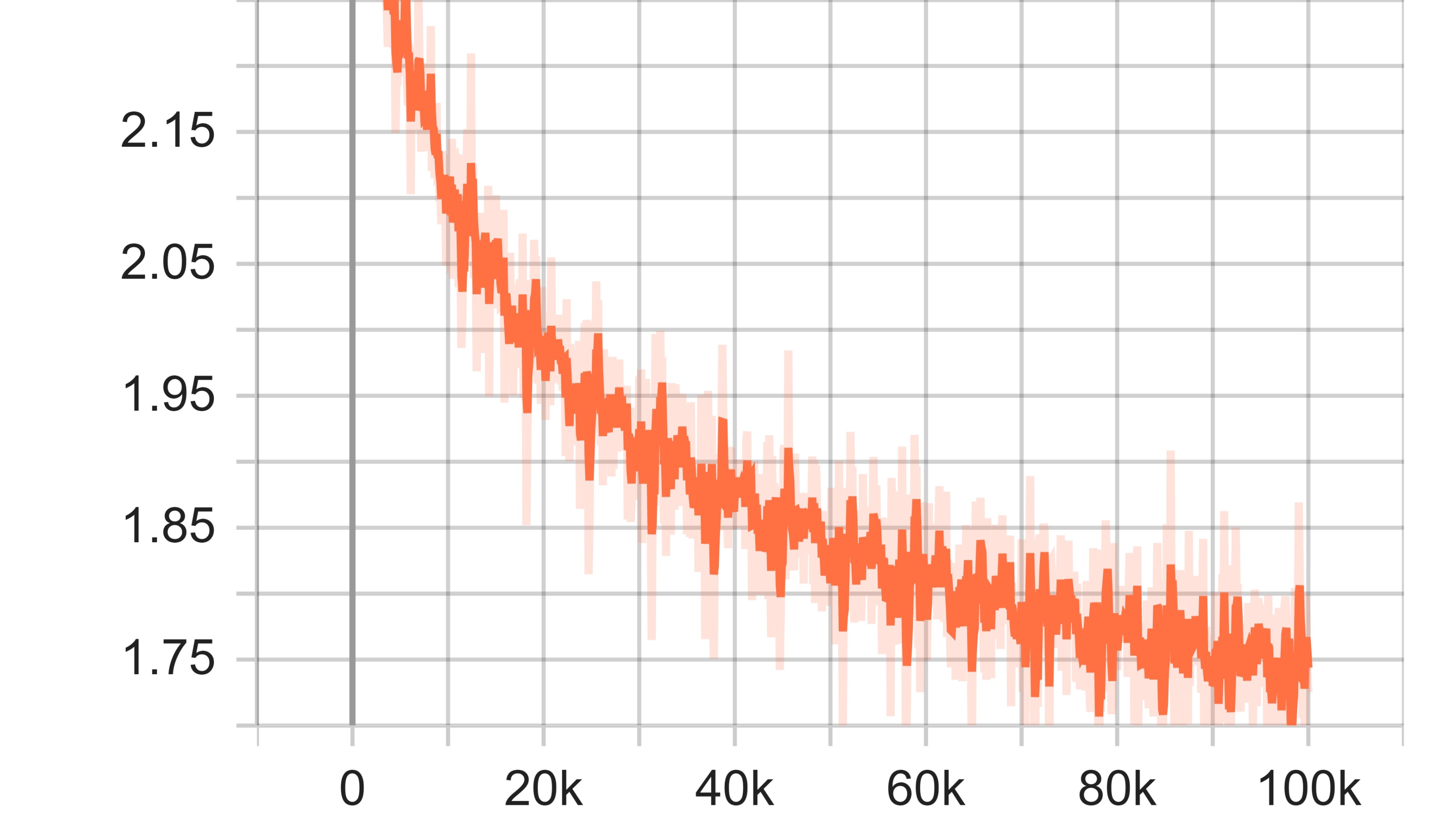}}
		\subfigure[Training loss on KdConv.]{
	\includegraphics[width=0.32\linewidth]{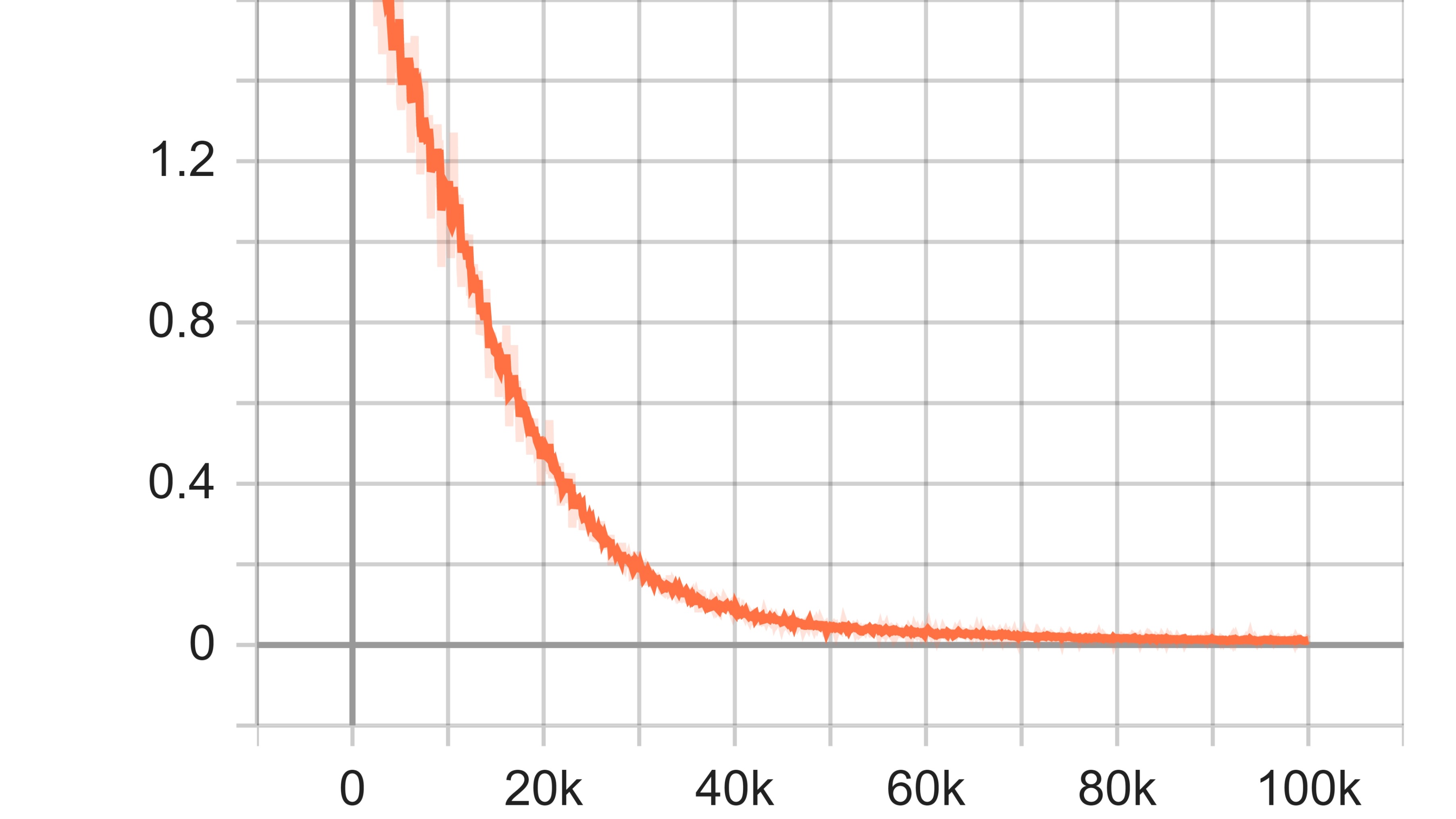}}
	\subfigure[Training loss on NLPCC.]{
	\includegraphics[width=0.32\linewidth]{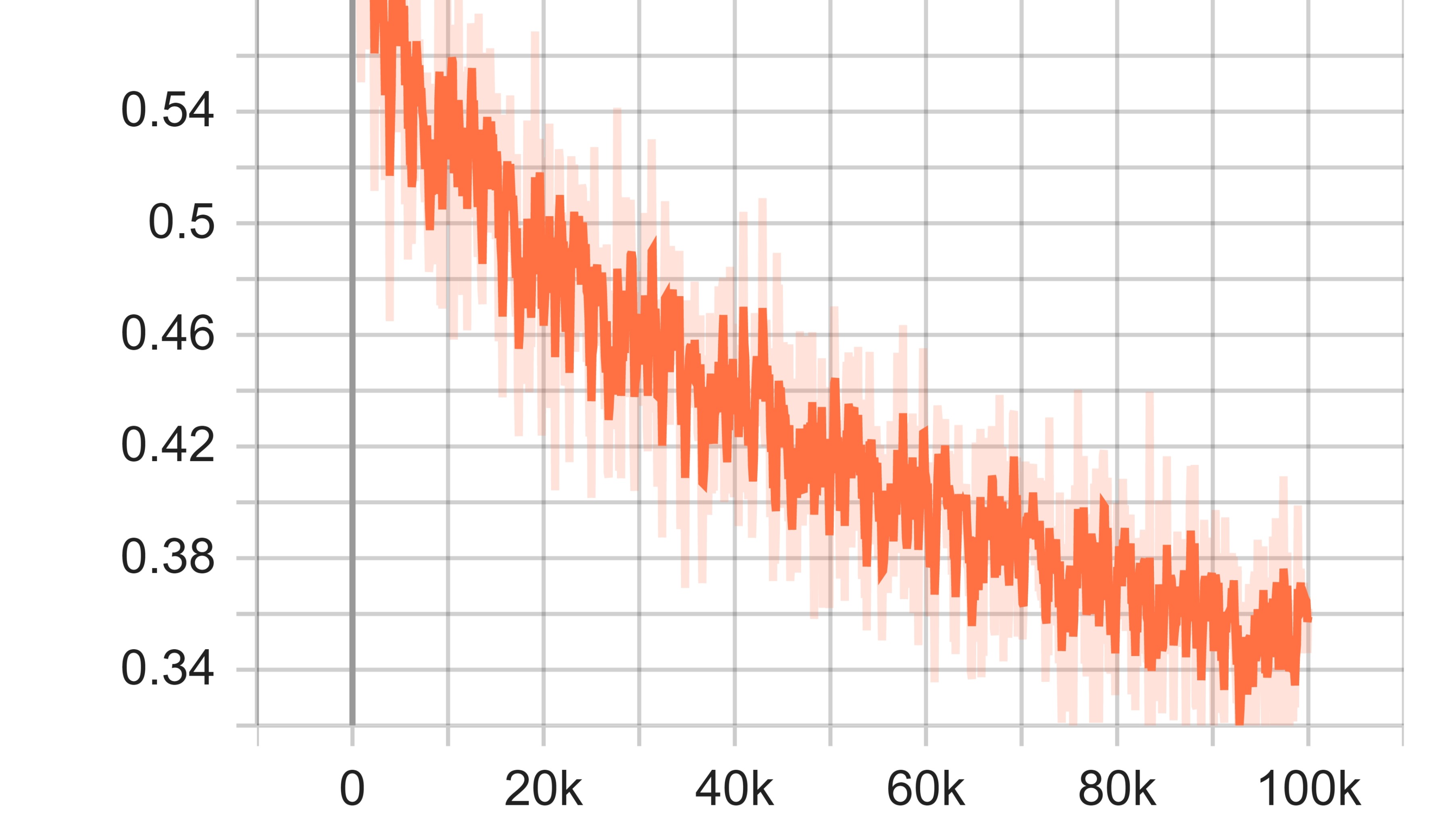}}

	\caption{The learning curves of tasks during multi-task training.}
	\label{fig:learning_curves}
\end{figure*}

We show the learning curves during multi-task training in Figure~\ref{fig:learning_curves}.

\end{document}